\documentclass[conference]{IEEEtran}
\IEEEoverridecommandlockouts
\usepackage{cite}
\usepackage{makecell}

\usepackage{amsmath,amssymb,amsfonts}
\usepackage{algorithmic}
\usepackage{graphicx}
\usepackage{textcomp}
\usepackage{xcolor}
\usepackage{hyperref}
\usepackage{graphicx}
\usepackage{multirow}
\usepackage{booktabs}
\usepackage{multirow}
\usepackage{bbding}
\usepackage{pifont}
\usepackage{wasysym}
\usepackage{amssymb}
\usepackage{footnote}
\usepackage{subfig}
\def\BibTeX{{\rm B\kern-.05em{\sc i\kern-.025em b}\kern-.08eml
    T\kern-.1667em\lower.7ex\hbox{E}\kern-.125emX}}
\begin{document}

\title{S-LAM3D: Segmentation-Guided Monocular 3D Object Detection via Feature Space Fusion \\
}

\author{\IEEEauthorblockN{Diana-Alexandra Sas}
\IEEEauthorblockA{\textit{Technical University of Cluj-Napoca} \\
\textit{Computer Science Department}\\
Cluj-Napoca, Romania \\
sas.cr.diana@student.utcluj.ro}
\and
\IEEEauthorblockN{Florin Oniga}
\IEEEauthorblockA{\textit{Technical University of Cluj-Napoca} \\
\textit{Computer Science Department}\\
Cluj-Napoca, Romania \\
florin.oniga@cs.utcluj.ro}
}

\maketitle 

\begin{abstract}
Monocular 3D Object Detection represents a challenging Computer Vision task due to the nature of the input used, which is a single 2D image, lacking in any depth cues and placing the depth estimation problem as an ill-posed one.
Existing solutions leverage the information extracted from the input by using Convolutional Neural Networks or Transformer architectures as feature extraction backbones, followed by specific detection heads for 3D parameters prediction. 
In this paper, we introduce a decoupled strategy based on injecting precomputed segmentation information priors and fusing them directly into the feature space for guiding the detection, without expanding the detection model or jointly learning the priors. The focus is on evaluating the impact of additional segmentation information on existing detection pipelines without adding additional prediction branches. 
The proposed method is evaluated on the KITTI 3D Object Detection Benchmark, outperforming the equivalent architecture that relies only on RGB image features for small objects in the scene: pedestrians and cyclists, and proving that understanding the input data can balance the need for additional sensors or training data. 

\end{abstract}

\section{Introduction}
3D Object Detection is a fundamental Computer Vision task, consisting in detecting and locating objects belonging to specific categories within a visual input. The difference between 2D and 3D object detection is the target environment - 2D or 3D space, and the predicted parameters which define the bounding boxes. For 3D object detection, the objects' 2D position and class are supplemented with their orientation relative to the coordinate system, this way defining complete 3D bounding boxes. The visual inputs used for solving this task range from single or multi-view images to RGB-D images, depth maps, voxel grids or point clouds. Depending on the type of input, the level of information cues available at training and inference time is different, Monocular 3D Object Detection remaining the most challenging approach to this visual task, due to its lack of depth cues.

One of the main focuses in solving the Monocular 3D Object Detection task is a proper choice for the feature extraction backbone, which could compensate to a certain level the lack of depth information from the input image. Convolutional Neural Networks (CNNs) were the most common choice as feature extraction backbones in the previous years, proving themselves highly performant (\!\!\cite{omni3d, kumar2022deviant, gupnetpp, lu2021geometry}). However, the increasing popularity of Transformer architectures opened the door for experimenting with different types of Vision Transformer architectures as backbones, for a better understanding of the long-range dependencies and the global context within the image (\!\!\cite{lam3d, dst3d}). One drawback of Transformer architectures is the need for bigger datasets for proper training, drawback that we try to surpass in this paper by further exploiting the information that can be extracted from the input data and using it for guiding the detection. 

To address the challenges posed by Monocular 3D Object Detection, multimodal detection pipelines have been studied, with the input being represented by multiple data modalities: RGB images complemented by other sensor data, such as LiDAR (\!\!\cite{lin2023mlfdet, Chen_2022_CVPR, clocs2020}) or radar (\!\!\cite{centerfusion}). Although this approach is suitable for sensor-rich environments, it usually involves separate encoders for each input modality and complex fusion techniques. A more lightweight and modular version of guiding the detection pipeline is injecting precomputed priors directly into the feature space. The priors, usually depth maps (\!\!\cite{zhang2023monodetr, msfnet3d, ding2020}), segmentation maps (\!\!\cite{monocinis}) or a combination of the two (\!\!\cite{vfmm3d}), are derived from the same input modality, the RGB image, and can enhance the spatial understanding without the need to expand the detection pipeline by additional prediction branches.

This paper proposes S-LAM3D, a framework for solving the Monocular 3D Object Detection task in a guided manner, built upon the following key contributions:
\begin{itemize}
    \item We use a vision foundation model to generate information priors and introduce a simple approach to inject these into a Monocular 3D Object Detection pipeline without joint training.
    \item We propose different fusion strategies and fusion points for emphasizing the relevant regions in the input image.
    \item Comprehensive experimental and ablation study with different fusion methods and points and their effect on guided Monocular 3D Object Detection
\end{itemize}

\section{Related Work}

\textbf{Vision Foundation Models.}
Recently, Foundation Models are revolutionizing the research world by their power to generalize to tasks and data distributions different to the ones seen during the training phase. Seen as a milestone in Computer Vision, Segment Anything Model (SAM) \cite{sam} is the first Vision Foundation Model, built to solve image segmentation and achieving an impressive zero-shot performance. The proposed architecture consists of an image encoder, a mask decoder and a flexible prompt encoder which works with two types of prompts: sparse (points, boxes) and dense (masks) \cite{sam}. Grounded SAM \cite{groundedsam} is introduced later, extending the capabilities of SAM by meeting the need of working with text prompts as well. The proposed approach is using Grounding DINO \cite {groundingdino}, an open-set 2D object detector that works with text inputs to generate the 2D bounding boxes and further use the bounding boxes as sparse prompts for SAM. 
Following the development of SAM, the authors of \cite{depthanything} proposed Depth Anything, a foundation model for monocular depth estimation. The model was jointly trained on large-scale labeled and unlabeled images for enhancing the data coverage \cite{depthanything} and further improved in \cite{depthanythingv2} by training on synthetic data and scaling up the capacity of the teacher model. There were improvements in the area of Monocular 3D Object Detection as well, by introducing DetAny3D \cite{detectanything}. The authors of \cite{detectanything} use pre-trained 2D foundation model SAM \cite{sam} for embedding the input RGB image and extracting pixel-level information, depth-pretrained DINO (\!\!\cite{unidepth}, \cite{dinov2}) for high-level geometric information and then fuse them together by a proposed 2D Aggregator which uses cross-attention layers \cite{detectanything}.

\textbf{Monocular 3D Object Detection.}
Monocular 3D Object Detection is a challenging task due to the nature of its input, a single RGB image. The goal is to infer the 3D location, orientation and dimensions of the objects of interest without any depth information. In order to tackle this problem, there are methods that rely on pseudo-LiDAR representations for estimating the depth parameter, as a way of imitating a LiDAR input (\!\!\!\cite{refinedmpl, Weng_2019_ICCV, vfmm3d}). Instead of using a pseudo-LiDAR representation, the authors from \cite{lu2021geometry} propose a Geometry Uncertainty Projection (GUP) module to handle this depth inference problem. Even though CNNs yield good performance, Transformer-based architectures started to gain popularity in Monocular 3D Object Detection as well, both at the decoder level (\!\!\cite{zhang2023monodetr, he2023ssdmonodetr, he2023s3monodetr}) and as feature extraction backbones (\!\!\cite{dst3d, lam3d}). However, state-of-the-art architectures for Monocular 3D Object Detection proved that only using the visual features from the input image falls short compared to methods that also benefit from depth information at training time, which lead to exploring guided mechanisms for the detection.

\textbf{Guided Monocular 3D Object Detection.}
As a way of compensating for this lack of depth cues, various authors have been experimenting with concurrently predicting depth maps and using them together with the visual features for guiding the regions of interest in the 3D detection pipeline. In paper \cite{ding2020}, the focus is on depth-guided convolutions in which the filters and their receptive fields are learned from the generated depth maps, as an alternative to generating a pseudo-LiDAR representation of the depth map. The authors from \cite{msfnet3d} propose a dual-branch network together with a pixel-level fusion of image and depth features. Zhang et al. introduce in \cite{zhang2023monodetr} a depth-guided Transformer which guides the detection process by contextual depth cues. The Transformer architecture consists of two encoders: visual and depth, so the input image is encoded from two perspectives: visual information and depth geometry. Following this pipeline, the depth-guided decoder leads to a better spatial understanding of the scene by capturing spatial cues from depth-guided regions in the image \cite{zhang2023monodetr}. 

Although not widely used, another manner to guide Monocular 3D Object Detection is by using segmentation maps. They don't provide the geometry captured by the depth maps, but still represent a good way of augmenting the visual features extracted from the input RGB image. Heylen et al. leverage in \cite{monocinis} an instance segmentation map as guidance for the detection pipeline, in which the instances of a target category are differentiated with different annotations. The instance segmentation maps can be obtained from any external network or ground truth instances \cite{monocinis}. With the increasing popularity of Vision Foundation Models, they started to be used as tools for generating information priors such as depth or segmentation maps. Ding et al. \cite{vfmm3d} use Depth Anything \cite{depthanything} to generate a depth map, Segment Anything \cite{sam} to generate an instance mask and further project them in a 3D space to convert the input RGB image into a pseudo-LiDAR representation on which they apply a 3D object detection pipeline.

\section{Method}
Our goal is to design and implement a solution for Monocular 3D Object Detection in the context of autonomous driving systems, which benefits from the input data at a higher level by not only extracting visual cues from the input image but also using an additional segmentation map to emphasize the regions of interest and guide the detection pipeline, without adding an extra prediction branch. The proposed approach extends our previous work, LAM3D \cite{lam3d}, a framework used for solving the task at hand with a Vision Transformer architecture as feature extraction backbone, by introducing a decoupled fusion strategy directly in the feature space. 

\begin{figure*}[!t]
    \centering
    \includegraphics[width=\textwidth]{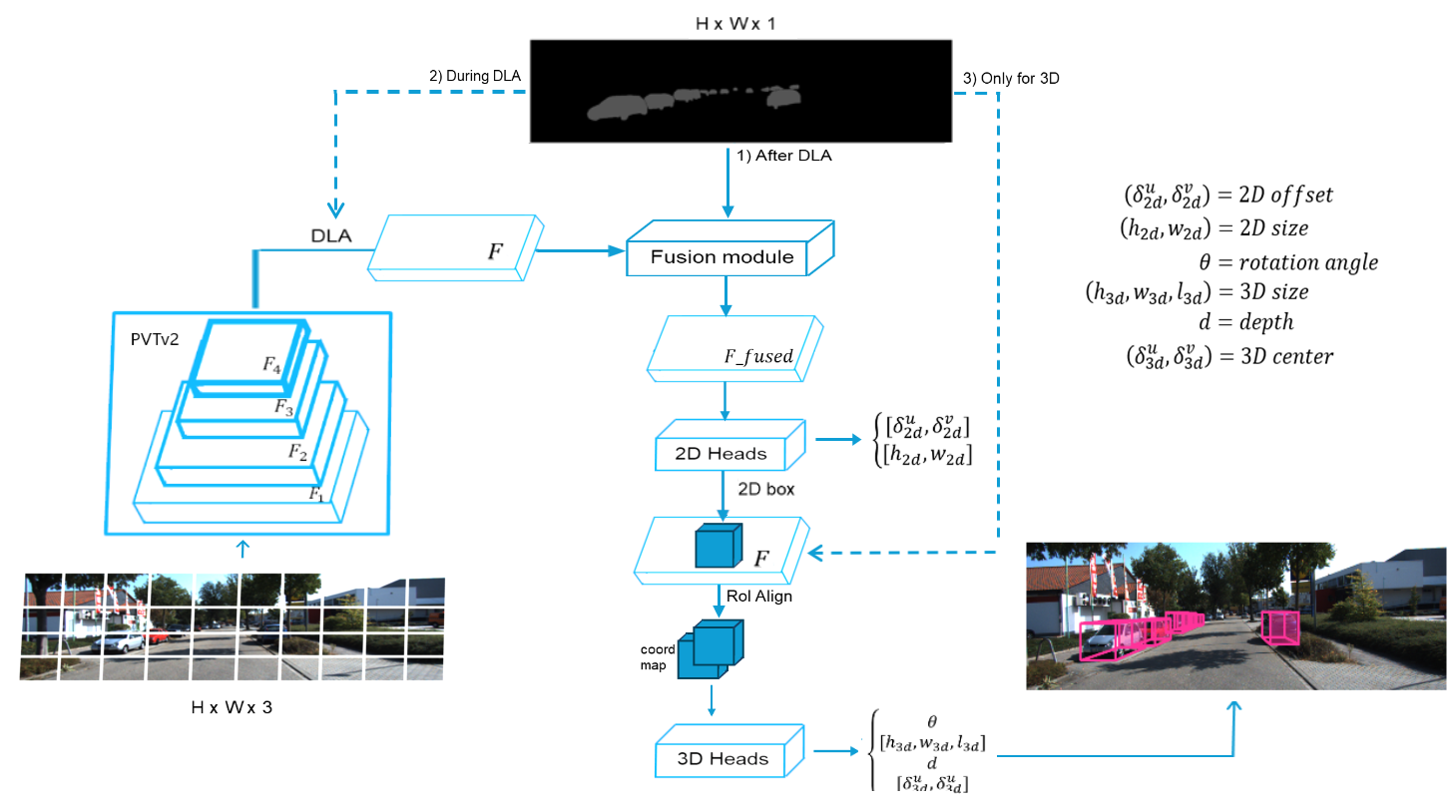}
    \caption{Architecture overview. The input is processed by a PVTv2 backbone \cite{Wang_2022} and the resulting feature maps are aggregated by using Deep Layer Aggregation (DLA) \cite{yu2019deep}. The different points where the segmentation priors are injected are represented with dashed lines. The best configuration is 1) After DLA in which the segmentation priors are injected after the aggregation module, fused together with the visual features and sent to the detection heads.}
    \label{fig:architecture}
\end{figure*}

The input is represented by a single 2D image and the additional information input. The 2D image is fed to a Transformer backbone for visual cues extraction and together with the information priors are sent to a simple fusion module which operates directly in the feature space. The fused feature map is used in the 3D detection pipeline. An overview of the architecture from \cite{lam3d} augmented with the information priors fusion strategies is presented in  Figure \ref{fig:architecture}.
This section presents the process of generating the information priors as well as the proposed fusion methods. 

\subsection{Information priors}
\begin{figure}
    \centering
    \includegraphics[width=1.0\linewidth]{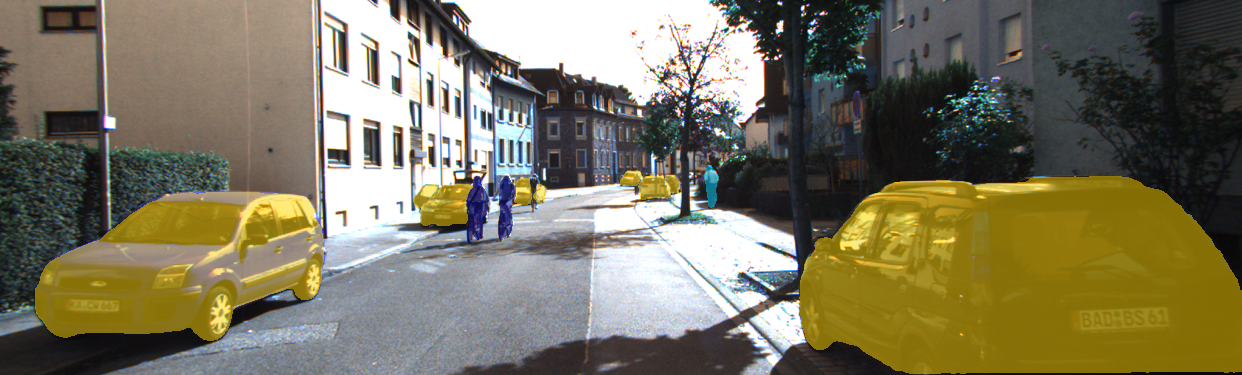}
    \caption{Semantic segmentation map overlay for a sample from KITTI 3D Object Detection Dataset \cite{KITTI}}
    \label{fig:overlay}
\end{figure}

\begin{figure}
    \centering
    \includegraphics[width=1.0\linewidth]{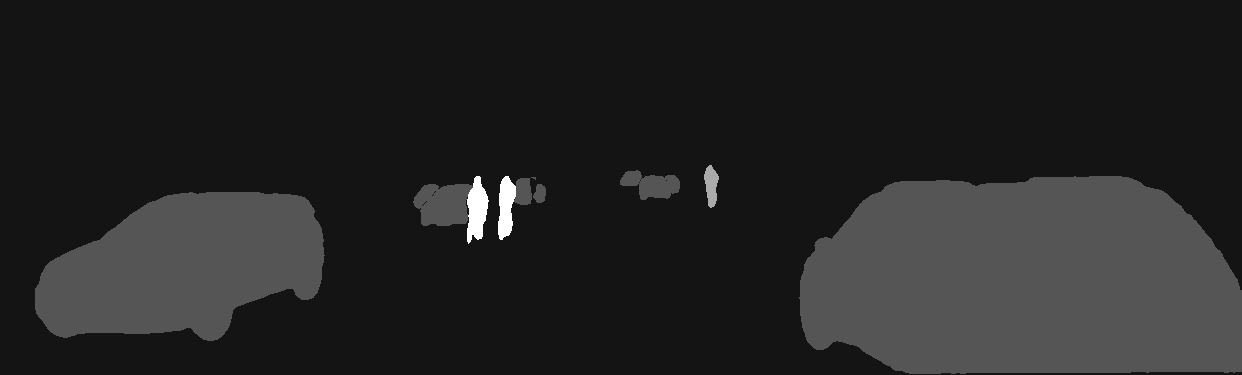}
    \caption{Grayscale semantic segmentation map generated for a sample from KITTI 3D Object Detection Dataset \cite{KITTI}}
    \label{fig:seg_map}
\end{figure}

The additional information injected in the detection pipeline is generated beforehand using Grounded SAM \cite{groundedsam}. The text prompts used for Grounding DINO \cite{groundingdino} are equivalent to the categories of interest for which we perform 3D Object Detection: car, pedestrian, cyclist. The 2D bounding boxes outputted by Grounding DINO are further used as spatial prompts for pretrained SAM \cite{sam} to generate the segmentation masks for the 2D detections. Such an inference result overlay is shown in Figure \ref{fig:overlay}. The resulting segmentation map is post-processed in a class-wise grayscale manner, with a different gray intensity for each category, as in Figure \ref{fig:seg_map}. The segmentation priors could be obtained with any other available segmentation network or existing ground truths, highlighting the flexible nature of our framework.

We use semantic segmentation maps as additional information in the experiments performed, but the proposed framework is universal enough to enable the use of any other information priors, such as depth maps. This flexibility is possible because we chose to decouple the injection mechanism from the actual prediction instead of adding additional prediction branches, resulting in a lightweight and universal framework. This design choice also allows to inject the information priors in different stages of the detection pipeline. 

\subsection{Fusion module}
The experiments we performed were focused around three fusion points: after the DLA module, when the visual features are already aggregated, before the aggregation module and only for the 3D detections heads. The fusion point with the best results is after the DLA module, the others are represented with dashed lines in Figure \ref{fig:architecture}.

The segmentation map information priors are spatially aligned with the input RGB image by applying identical transformations to both inputs: random flipping and cropping as data augmentation, followed by an affine transformation to the target resolution using billinear interpolation. This is a pre-fusion step which ensures perfect pixel-wise alignment between the input RGB image and the information priors. The segmentation map is bilinearly upsampled to match the spatial dimensions of the feature map before fusion. Feature standardization across the mean and variance is performed for both the segmentation map and the aggregated features in order to have a comparable range with similar statistical properties for a more robust fusion process and improved training stability. The standardized features are combined based on an element-wise multiplicative fusion approach which allows the segmentation map to modulate the extracted visual features in order to emphasize the regions of interest while suppressing irrelevant background areas. This way, the segmentation map guides the network to focus only on object-relevant features, acting as an attention mechanism. Based on our experiments, element-wise multiplication achieved better results than concatenation or attention-based fusion.

The spatial structure for both inputs is preserved and the features are stronger in regions where both the RGB input extracted features and the segmentation map are highly confident. The final convolutional layer projects the fused features to a 64-channel feature space suitable for the detection heads. This representation is used for the 2D parameters prediction, depth estimation and 3D bounding box regression.

\section{Experiments}

\subsection{Setup}
\subsubsection{Dataset}
The dataset used for training and evaluation is KITTI 3D Object Detection \cite{KITTI} that comprises 7481 training and 7518 testing images. Each labeled bounding box is marked as visible, semi-occluded, fully occluded or truncated for better understanding of the edge cases and better evaluation. In this paper, the experiments were conducted on the common validation split \cite{lu2021geometry} in which the training data is further divided into a training set of 3712 images and a validation set of 3769 images, as the ground truths for the official testing set are not made publicly available.

\subsubsection{Evaluation protocol}
The 3D Object Detection performance is evaluated using the PASCAL criteria, also used for 2D Object Detection. The 3D bounding box overlap is set to 70\% for cars and 50\% for pedestrians and cyclists. As an additional configuration we also evaluate a 3D bounding box overlap of 50\% for cars and 30\% for pedestrians and cyclists. The difficulty levels are defined as in the KITTI 3D Object Detection Benchmark (Easy, Moderate, Hard), with different thresholds for minimum bounding box height, occlusion level and truncation \cite{KITTI}.

\subsubsection{Implementation details}
The segmentation maps were generated in advance using Grounded SAM \cite{groundedsam}. Both the input RGB image and the segmentation map have a resolution of 384 x 1280. The proposed model was trained on 2 NVIDIA Tesla V100 GPUs with batch size 12 for 140 epochs. Adam optimizer was used for training with an initial learning rate of $1.25 \times 10^{-3}$ and a rate decay set to $0.1$ at epochs \{90,120\}.

\subsection{Results}
We report the results on the validation subset of KITTI 3D Object Detection Dataset and compare them with our baseline method, LAM3D \cite{lam3d}, which doesn't use any kind of information priors. The results in Table \ref{table:main_results} highlight the improvement in performance that our segmentation-guided method brings in the detection of small and often-overlooked object categories: pedestrians and cyclists. These categories are underrepresented in the training dataset and hard-to-detect targets due to their smaller size and appearance. For pedestrians, there are substantial gains for our proposed segmentation-guided method, S-LAM3D, of 0.94\% at $AP_{70}$ at Hard difficulty, 0.91\% at Mod. difficulty and 1.66\% at Easy difficulty. For the cyclist category the improvement at $AP_{70}$ is of 0.36\% for Hard, 0.48\% for Mod. and 0.08\% for Easy. There is a slight drop in car $AP_{3D}$ compared to the baseline, which we attribute to the quality of the segmentation priors and the fusion technique. When using precomputed information priors and multiplicative fusion there is a risk of not having segmentation masks for all ground-truth instances, this way suppressing relevant regions from the input. By choosing to emphasize the regions with valid segmentation masks, the network is focusing on spatially accurate predictions, but it can miss detections with lower quality, which also explains the performance for $IOU=0.5$.

\label{cap:main_results}
\begin{table}[ht!]
\caption{Comparison of Architectures on KITTI Val}
\centering
\resizebox{\columnwidth}{!}{%
\begin{tabular}{@{}lllllccccccc@{}}
\toprule
\multirow{2}{*}{Method}  & \multirow{2}{*}{Category} & \multicolumn{3}{c}{$AP_{3D}(IOU=0.7|R_{40})$} & \multicolumn{3}{c}{$AP_{3D}(IOU=0.5|R_{40})$} \\ \cmidrule(lr){3-5} \cmidrule(lr){6-8}
 &  &  Easy & Mod. & Hard & Easy & Mod. & Hard \\ \midrule
   & Car & \textbf{22.55} & \textbf{15.66} & \textbf{12.85} & \textbf{60.08} & \textbf{45.62} & \textbf{39.50} \\ 
{LAM3D \cite{lam3d}}   & Pedestrian & {8.98} & {6.64} & {5.33} & \textbf{26.44} & {20.81} & \textbf{17.07} \\ 
   & Cyclist & {8.09} & {3.87} & {3.78} & \textbf{21.48} & {11.19} & \textbf{11.01} \\ \midrule
& Car & {19.32} & {13.62} & {11.69} & {58.32} & {43.10} & {37.13} \\ 
\textbf{S-LAM3D (Ours)}   & Pedestrian & \textbf{10.64} & \textbf{7.58} & \textbf{6.24} & {26.40} & \textbf{20.82} & {16.96} \\ 
   & Cyclist & \textbf{8.17} & \textbf{4.35} & \textbf{4.14} & {20.95} & \textbf{12.08} & {10.96} \\ \bottomrule
\end{tabular}%
}
\label{table:main_results}      
\end{table}

Despite the slight drop in AP3D for cars, the predictions variance values are lower across all difficulty levels and thresholds (Table \ref{table:var}). This indicates a more robust and confident network that still benefits from the segmentation priors.  

\begin{table}[ht!]
\centering
\caption{Variance values of the 3D predictions for the \textit{car} category on KITTI Val}
\resizebox{\columnwidth}{!}{%
\begin{tabular}{@{}ll*{6}{c}@{}} 
\toprule
\multirow{2}{*}{Method} & \multirow{2}{*}{Input}
& \multicolumn{3}{c}{$\sigma_{3D}(\mathrm{IoU}=0.7 \mid R_{40})$}
& \multicolumn{3}{c}{$\sigma_{3D}(\mathrm{IoU}=0.5 \mid R_{40})$} \\
\cmidrule(lr){3-5} \cmidrule(lr){6-8}
 &  & Easy & Mod. & Hard & Easy & Mod. & Hard \\
\midrule
LAM3D \cite{lam3d} & RGB Image
& 5.73 & 5.60 & 4.86 & 14.27 & 16.44 & 16.06 \\
S\text{-}LAM3D (Ours) &
\raisebox{-0.35\baselineskip}{%
  \begin{minipage}[t]{2.5cm}\raggedright
    RGB Image\\
    + Segmentation Map
  \end{minipage}%
}
& \textbf{5.61} & \textbf{5.44} & \textbf{4.61} & \textbf{14.25} & \textbf{16.42} & \textbf{16.04} \\
\bottomrule
\end{tabular}%
}
\label{table:var}
\end{table}

Table \ref{table:test_kitti} shows that S-LAM3D outperforms MonoCInIS \cite{monocinis}, another segmentation-guided Monocular 3D Object Detection solution, in $AP_{3D}$ and $AP_{BEV}$ metrics for car category on KITTI test set, evaluation performed on KITTI servers with withheld ground-truths.  
A significant part of the improvement can be attributed to the chosen backbone and architecture rather than segmentation alone. However, segmentation priors, while not outperforming the RGB-only car detection, improve the stability and robustness of the detections. S-LAM3D reports evaluation results on all three categories: car, pedestrian and cyclist, showing consistent improvements for the other two categories, which are typically more challenging in a monocular context.
\begin{table}[ht!]
\caption{Comparison of Segmentation-guided Architectures on KITTI Test cars}

\centering
\resizebox{\columnwidth}{!}{%
\begin{tabular}{@{}llcccccc@{}}
\toprule
\multirow{2}{*}{Method}  & \multirow{2}{*}{Backbone} & \multicolumn{3}{c}{$AP_{3D}(IOU=0.7|R_{40})$} &  \multicolumn{3}{c}{$AP_{BEV}(IOU=0.7|R_{40})$} \\ \cmidrule(lr){3-5} \cmidrule(lr){6-8}
 &  &  Easy & Mod. & Hard & Easy & Mod. & Hard \\ \midrule
{MonoCInIS \cite{monocinis} }   & ResNet101 & 15.82 & 7.94 & 6.68 & 22.28 & 11.64 & 9.95 \\ \midrule
\textbf{S-LAM3D (Ours)}   & PVTv2 b2 & \textbf{18.97} & \textbf{12.66} & \textbf{10.76} & \textbf{27.25} & \textbf{18.70} & \textbf{15.77} \\ \bottomrule
\end{tabular}%
}
\label{table:test_kitti} 

\end{table}

\subsection{Ablation Study}
In order to evaluate the influence that injecting segmentation priors has on solving the Monocular 3D Object Detection task, several experiments were conducted on the KITTI 3D Object Detection validation split and will be presented in this section.

\subsubsection{Fusion techniques}
Different fusion strategies for the visual features and segmentation priors were validated and the results reported in Table \ref{table:diff_fusion_strategies}. Best results were achieved by performing a multiplicative fusion, as it works as a lightweight attention mechanism, without introducing extra parameters or complex interactions. Concatenation lacks a strong interaction between the two types of features and fully leaves the network to merge the features.

\begin{table}[ht!]
\caption{Different fusion techniques. Results are reported on KITTI Val.}
\centering
\resizebox{\columnwidth}{!}{%
\begin{tabular}{@{}cllllccccccc@{}}
\toprule
\multirow{2}{*}{Fusion technique}  & \multirow{2}{*}{Category} & \multicolumn{3}{c}{$AP_{3D}(IOU=0.7|R_{40})$} & \multicolumn{3}{c}{$AP_{3D}(IOU=0.5|R_{40})$} \\ \cmidrule(lr){3-5} \cmidrule(lr){6-8}
 &  &  Easy & Mod. & Hard & Easy & Mod. & Hard \\ \midrule
& Car & {18.06} & {12.99} & {10.61} & {56.60} & {41.44} & {36.74}  \\
Concatenation & Pedestrian & {8.38} & {6.22} & {4.95} & {23.05} & {17.40} & {14.05} \\
& Cyclist & \textbf{8.44} & \textbf{4.69} & {4.04} & {20.36} & {11.80} & {10.52}   \\ \hline
& Car & {19.15} & \textbf{13.70} & {11.27} & {57.55} & {41.73} & {35.65}  \\
Attention & Pedestrian & {8.28} & {6.17} & {4.99} & {25.92} & {20.55} & {16.75} \\
& Cyclist & {7.67} & {3.84} & {3.69} & {20.34} & {11.08} & {9.85}  \\ \hline
& Car & \textbf{19.32} & {13.62} & \textbf{11.69} & \textbf{58.32} & \textbf{43.10} & \textbf{37.13} \\ 
\textbf{Multiplication}   & Pedestrian & \textbf{10.64} & \textbf{7.58} & \textbf{6.24} & \textbf{26.40} & \textbf{20.82} & \textbf{16.96} \\ 
   & Cyclist & {8.17} & {4.35} & \textbf{4.14} & \textbf{20.95} & \textbf{12.08} & \textbf{10.96} \\ \bottomrule
\end{tabular}%
}
\label{table:diff_fusion_strategies}              
\end{table}

\subsubsection{Fusion points}
Different fusion points were proposed and also highlighted in the architecture overview in Figure \ref{fig:architecture}. Injecting the segmentation priors after the aggregation of the multi-scale features is maximizing the impact on the spatial reasoning, compared to only injecting them before the 3D detection heads, where their effect is limited. Another proposed fusion point was during the DLA stage, but this interferes with the hierarchical feature learning process. The results for all fusion points are shown in Table \ref{table:diff_fusion_points}.   

\begin{table}[ht!]
\caption{Different fusion points. Results are reported on KITTI Val.}
\centering
\resizebox{\columnwidth}{!}{%
\begin{tabular}{@{}cllllccccccc@{}}
\toprule
\multirow{2}{*}{Fusion point}  & \multirow{2}{*}{Category} & \multicolumn{3}{c}{$AP_{3D}(IOU=0.7|R_{40})$} & \multicolumn{3}{c}{$AP_{3D}(IOU=0.5|R_{40})$} \\ \cmidrule(lr){3-5} \cmidrule(lr){6-8}
 &  &  Easy & Mod. & Hard & Easy & Mod. & Hard \\ \midrule
& Car & \textbf{19.56} & 13.59 & \textbf{11.88} & 57.72 & 42.67 & 36.61 \\
During DLA & Pedestrian & 7.89 & 6.30 & 5.04 & 23.06 & 17.99 & 14.66 \\
& Cyclist & 7.10 & 4.04 & 3.42 & \textbf{21.32} & \textbf{12.12} & 10.82  \\ \hline
& Car & 17.21 & 12.91 & 11.35 & 58.27 & \textbf{43.28} & 37.12 \\
Only for 3D  & Pedestrian & 8.48 & 6.40 & 5.02 & 24.49 & 19.63 & 15.95 \\ 
   & Cyclist & 5.77 & 3.01 & 2.53 & 17.87 & 9.63 & 8.60 \\ \hline
& Car & {19.32} & \textbf{13.62} & 11.69 & \textbf{58.32} & 43.10 & \textbf{37.13} \\ 
\textbf{After DLA} & Pedestrian & \textbf{10.64} & \textbf{7.58} & \textbf{6.24} & \textbf{26.40} & \textbf{20.82} & \textbf{16.96} \\ 
   & Cyclist & \textbf{8.17} & \textbf{4.35} & \textbf{4.14} & 20.95 & 12.08 & \textbf{10.96} \\ \bottomrule
\end{tabular}%
}
\label{table:diff_fusion_points}              
\end{table}

\subsubsection{Segmentation priors generation}
In order to demonstrate the modularity of our framework and the importance of high-quality segmentation priors we evaluate the performance of the same architecture, S-LAM3D, with priors generated by different segmentation networks. Grounded SAM \cite{groundedsam} yields better results than DeepLabv3+ \cite{deeplabv3} due to its powerful zero-shot generalization, producing more accurate segmentation masks. 
The results are in Table \ref{table:diff_seg_net}.

\begin{table}[ht!]
\caption{Different segmentation networks. Results are reported on KITTI Val.}
\centering
\resizebox{\columnwidth}{!}{%
\begin{tabular}{@{}cllllccccccc@{}}
\toprule
\multirow{2}{*}{Segmentation network}  & \multirow{2}{*}{Category} & \multicolumn{3}{c}{$AP_{3D}(IOU=0.7|R_{40})$} & \multicolumn{3}{c}{$AP_{3D}(IOU=0.5|R_{40})$} \\ \cmidrule(lr){3-5} \cmidrule(lr){6-8}
 &  &  Easy & Mod. & Hard & Easy & Mod. & Hard \\ \midrule
& Car & 18.86 & 12.88 & 10.58 & \textbf{59.80} & 41.41 & 35.45 \\
DeepLabv3+ \cite{deeplabv3} & Pedestrian & 8.31 & 6.65 & 5.22 & 25.05 & 18.74 & 15.13 \\
& Cyclist & 5.98 & 3.39 & 2.65 & \textbf{23.93} & \textbf{13.06} & \textbf{11.53}  \\ \hline
& Car & \textbf{19.32} & \textbf{13.62} & \textbf{11.69} & 58.32 & \textbf{43.10} & \textbf{37.13} \\ 
Grounded SAM \cite{groundedsam} & Pedestrian & \textbf{10.64} & \textbf{7.58} & \textbf{6.24} & \textbf{26.40} & \textbf{20.82} & \textbf{16.96} \\ 
   & Cyclist & \textbf{8.17} & \textbf{4.35} & \textbf{4.14} & 20.95 & 12.08 & 10.96 \\ \bottomrule
\end{tabular}%
}
\label{table:diff_seg_net}              
\end{table}


\subsection{Computational analysis}
The inference running time of the proposed method is on average 68 ms/image on KITTI 3D Dataset, with an increase of 5 ms compared to the baseline LAM3D \cite {lam3d}. The memory usage is up to 5.2GB, compared to 4.3GB without the decoupled injection mechanism and fusion module. These parameters prove that the proposed method adds an insignificant overload in terms of inference time and memory, while bringing a meaningful performance improvement for small objects.

\section{Qualitative Results}
\begin{figure*}[htbp!]
    \centering
    \subfloat[LAM3D \cite{lam3d}]{\includegraphics[width=0.45\textwidth]{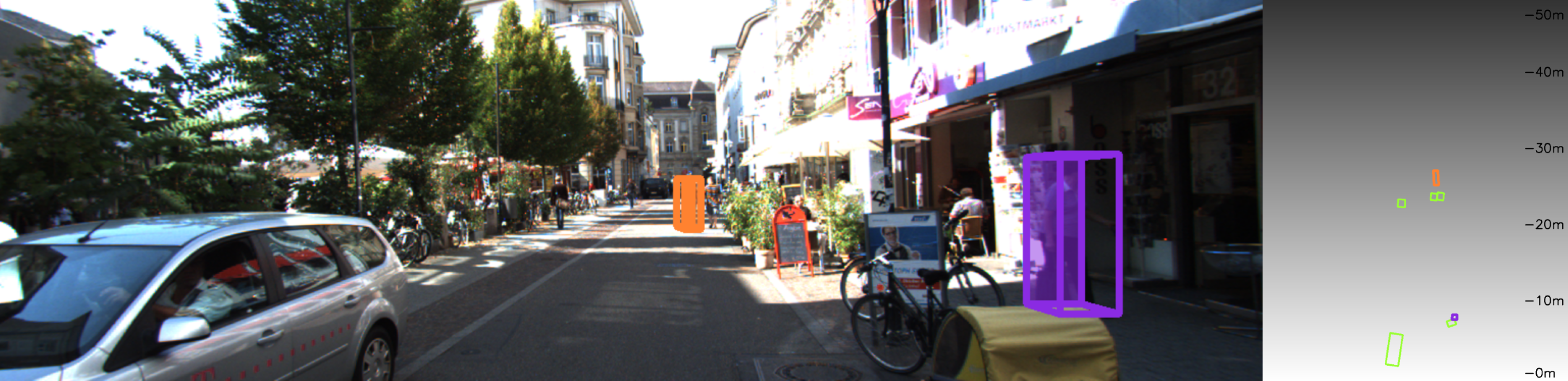}\label{fig:15}}
    \hspace{0.02\textwidth}
    \subfloat[S-LAM3D (Ours)]{\includegraphics[width=0.45\textwidth]{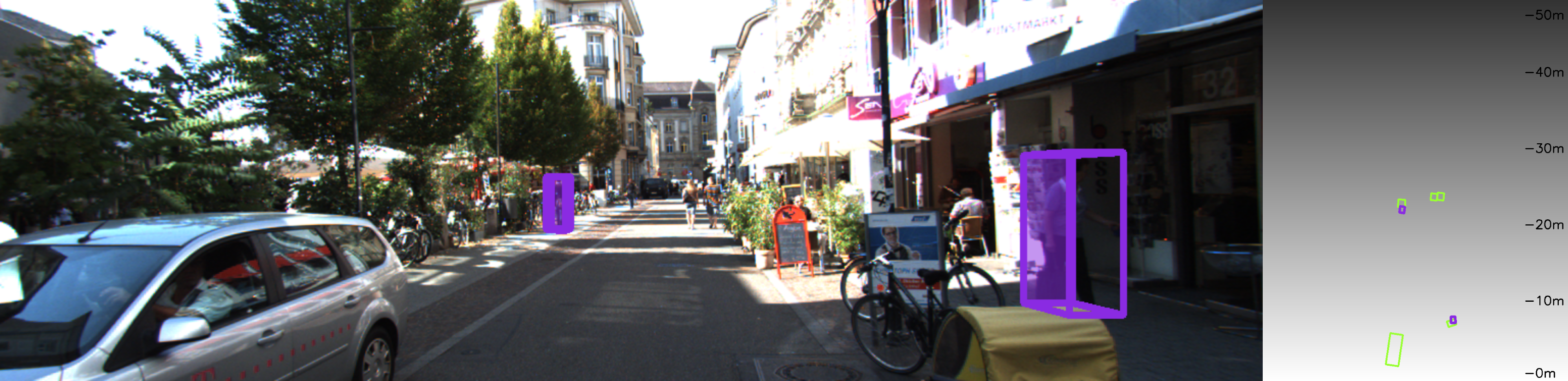}\label{fig:15s}} \\
    \subfloat[Segmentation overlay]{\includegraphics[width=0.45\textwidth]{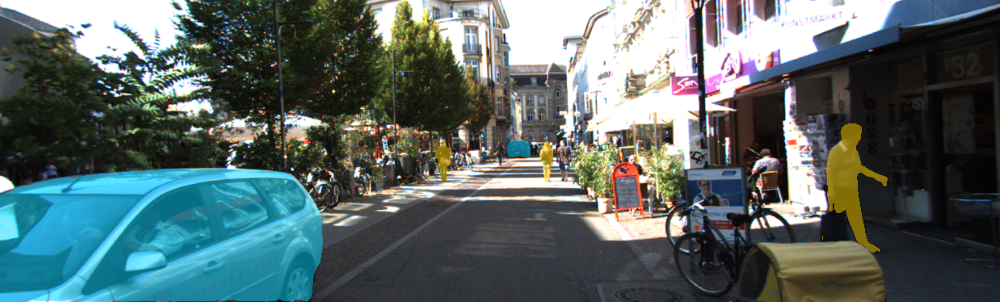}\label{fig:15_seg}}
    \caption{Qualitative results for the proposed method. The ground truth is represented by green bounding boxes, the purple bounding boxes represent pedestrians and the orange bounding boxes cyclists.}
    \label{fig:qualitative_out}
\end{figure*}

In Figure \ref{fig:qualitative_out} we present qualitative results for the proposed method, S-LAM3D, comparing to the baseline method LAM3D \cite{lam3d}. A more accurate detection for small objects can be observed when using segmentation priors.

\section{Conclusion}
In this paper, we propose S-LAM3D, a novel framework that uses a decoupled strategy for injecting segmentation priors into the architecture and a light fusion module for combining the visual features with the segmentation priors for solving the Monocular 3D Object Detection task. The extensive experiments conducted show how the performance is influenced by different fusion  strategies or points and by the quality of the priors. The proposed method brings a significant performance improvement for small objects in the scene, belonging to pedestrian and cyclist categories, proving how properly modulating the visual features with the segmentation priors leads to a better 3D detection in the monocular context.

\section*{Acknowledgment}
This work was supported in part by the project ”Romanian Hub for Artificial Intelligence-HRIA”, Smart Growth, Digitization and Financial Instruments Program, MySMIS no. 334906. Conference registration and participation costs were supported by the Technical University of Cluj-Napoca, Romania.

\bibliographystyle{ieeetr}
\bibliography{ref.bib}

\end{document}